\documentclass{article}

\PassOptionsToPackage{numbers}{natbib}
\usepackage{amsmath}
\usepackage{enumitem}
\usepackage{amssymb}
\usepackage{multirow}
\usepackage{tabularx}
\usepackage{dsfont}
\usepackage[ruled]{algorithm2e}
\usepackage[final]{nips_2017}

\usepackage[utf8]{inputenc} 
\usepackage[T1]{fontenc}    
\usepackage{hyperref}       
\usepackage{url}            
\usepackage{booktabs}       
\usepackage{amsfonts}       
\usepackage{nicefrac}       
\usepackage{microtype}      

\title{End-to-End Offline Goal-Oriented Dialog Policy Learning via Policy Gradient}

\author{
   Li Zhou \\
   Amazon \\
   Seattle, WA \\
   \texttt{lizhouml@amazon.com} \\
   \And
   Kevin Small \\
   Amazon \\
   Seattle, WA \\
   \texttt{smakevin@amazon.com} \\
   \AND
   Oleg Rokhlenko \\
   Amazon \\
   Seattle, WA \\
   \texttt{olegro@amazon.com} \\
   \And
   Charles Elkan \\
   Amazon \\
   Seattle, WA \\
   \texttt{elkanc@amazon.com} \\
}

\begin{document}

\maketitle

\begin{abstract}
Learning a goal-oriented dialog policy is generally performed offline with
supervised learning algorithms or online with reinforcement learning (RL).
Additionally, as companies accumulate massive quantities of dialog transcripts
between customers and trained human agents, encoder-decoder methods have gained
popularity as agent utterances can be directly treated as supervision without
the need for utterance-level annotations. However, one potential drawback of
such approaches is that they myopically generate the next agent utterance
without regard for {\em dialog-level} considerations. To resolve this concern,
this paper describes an {\em offline} RL method for learning from {\em
  unannotated} corpora that can optimize a goal-oriented policy at both the
utterance and dialog level. We introduce a novel reward function and use both
on-policy and off-policy policy gradient to learn a policy offline without
requiring online user interaction or an explicit state space definition.
\end{abstract}

\section{Introduction}

Companies are increasingly interested in building goal-oriented dialog
systems for domains such as customer service and reservation systems. One
difference between building chatbots in industry and academia is that
companies usually possess a vast amount of dialogs where 
conversation transcripts are between customers and {\em trained} human agents. For example, customer service live chat systems are usually operated by human agents, and these agents have to take weeks of training courses before being deemed qualified to handle customer issues. Thus, agent utterances in these dialogs can be treated as actions produced by a near-optimal policy for machine learning algorithms. To build industry-quality goal-oriented chatbots, one important direction of research is effectively utilizing these corpora of \textit{trained agent-customer transcripts (TACTs)} to learn dialog policies offline while minimizing the dependence on domain knowledge and corresponding human effort.

Recently, several supervised learning algorithms have been
proposed to learn goal-oriented dialog policies~\cite{bordes2016learning,P17-1062}. In these algorithms, the agent
utterances are treated as labels, and the models are trained to maximize the
likelihood of agent utterances. While this is intuitive, supervised learning
models are trained to only optimize myopic rewards (i.e., likelihood of the next
utterances instead of the goal of the whole dialog). Simply imitating the agent
utterances in the training data is problematic and can suffer from compounding
errors in multi-turn dialogs~\cite{ross2011reduction}.

Reinforcement learning (RL), on the other hand, is a natural choice to learn
goal-oriented dialog policies as it optimizes long-term cumulative
reward and selects the best action to achieve the goal given the current
state. Recently, many RL-based algorithms have been proposed and achieved good
results~\cite{peng2017composite,dhingra2017,young2013pomdp,su2016online}. However, there are
two main drawbacks when directly applying reinforcement learning to dialog
policy learning: (1) RL algorithms for learning dialog managers are generally trained online, requiring interaction with real humans or user simulators during training and are not naturally suited to fully utilize large TACT corpora,
  (2) Most proposed algorithms require pre-defining a set of discrete actions and
slots for the state and action representation, which requires significant human
effort and domain knowledge -- thus, not readily generalizing to new domains.

To overcome the first problem, one can pre-train RL models offline with supervised learning methods and then fine-tune policies online with real humans (or a user simulator) in the loop~\cite{su2016continuously,P17-1062,su2017sample}. In addition to still requiring human interaction, pre-training also requires human effort to annotate each agent utterance in the training data with a pre-defined action and annotate entities with pre-defined slots (i.e., they require annotated data). \citet{kandasamy2016batch} proposed a batch policy gradient algorithm, which weighs trajectories in the dataset by importance sampling. Thus, it can learn
dialog policy offline, but may suffer from large gradient variance due to importance
sampling. Secondly, this algorithm also requires that reward signals exist
in the dataset, which is not necessarily the case in practice.

To overcome the problem of requiring a pre-defined action space, several works~ \cite{Serban2016,eric2017copy,li2016deep} proposed using an encoder-decoder (i.e., sequence-to-sequence) neural architecture to learn a model for directly generating dialog responses. These approaches enable end-to-end learning and do not need domain knowledge to define the action or state space such that they can easily generalize to different domains. However, similar to the previously mentioned supervised learning algorithms, these algorithms myopically optimize the likelihood of next utterance and neglect the overarching goal of the dialog. They
also tends to generate very generic responses~\cite{kandasamy2016batch,li2016deep}.

Inspired by recent RL-based algorithms for sequence
prediction~\cite{Bahdanau2017an,ranzato2015sequence}, we propose a RL-based
end-to-end dialog policy learning algorithm that can overcome these two drawbacks. Specifically, we model agent response generation
as a Markov decision process (MDP) in which each episode (from an initial state
to a terminal state) corresponds to a sequence of words in an agent
utterance. Note that this is different from the MDP defined in related
literature~\cite{dhingra2017,su2017sample}, where each episode corresponds to a sequence of dialog acts.
This MDP has a known transition function and defined reward function, so we can
learn the optimal policy offline with an on-policy algorithm without interacting
with real users.
We adopt an neural encoder-decoder architecture to parameterize the policy such
that our algorithm can be trained on unannotated data without defining dialog
acts or slots, and still reap the benefits of the latest encoder-decoder architectures.
The main contributions of this paper are as follows:
\begin{enumerate}[leftmargin=0.5cm]
  \item We propose a novel reward function that takes into account both
    utterance-level and dialog-level rewards. This reward function guides the
    algorithm to optimize not only the next agent utterances, but also the
    overall dialog goals.
  \item When learning the optimal policy of the defined MDP, we propose to use
    off-policy policy gradient to accelerate the convergence of on-policy
    policy gradient. 
  \item Our algorithm achieves better results than state-of-the-art
    sequence-to-sequence models on bAbI task
    6,\footnote{https://github.com/facebook/bAbI-tasks} a widely used
    goal-oriented dialog dataset. 
  \item The proposed algorithm learns dialog policy from unannotated dataset
    offline without interacting with real users. This enables us to utilize the vast
    amount of existing dialogs in TACTs, and makes building
    industry-quality chatbots possible.
\end{enumerate}

\section{Related Work}

There are two main approaches to learning a dialog policy in the literature, action prediction (i.e., utterance-level prediction) and sequence prediction (i.e., token-level prediction).

\textbf{Dialog policy learning as action prediction.} In this line of work, the
algorithms first predict the agent's next action, and then generate agent
utterances based on the action. Some supervised learning algorithms treat each
action as one class, and then perform multi-class classification. For example,
\citet{bordes2016learning} and \citet{P17-1062} proposed to use Memory Network
and recurrent neural network (RNN) as the underlying multi-class classification
model respectively. Usually, these algorithms are followed by training with
reinforcement learning to further refine the
model~\cite{su2016continuously,P17-1062,su2017sample,williams2016end}. \citet{P17-1062} achieved the state-of-the-art
on bAbI dialog dataset with a neural architecture, by relying on injecting
domain-specific knowledge and constraints, such as action masks. Meanwhile, in
this paper, we focus on learning dialog policy without such domain
knowledge. Many RL algorithms have also been
proposed~\cite{peng2017composite,dhingra2017,young2013pomdp,su2016online}. Usually,
these algorithms need to interact with real users or user simulators, which makes
them difficult to scale to industry-quality applications. Some batch reinforcement
learning algorithms have been
proposed~\cite{pietquin2011sample,li2014temporal}. However, these algorithms
requires annotated agent actions and rewards, which are often not available.

\textbf{Dialog Policy Learning as sequence prediction.} The algorithms in this
line of work generate agent utterances token by token given the dialog context~\cite{Serban2016,eric2017copy,sordoni2015a}. Usually these algorithms adopt
an encoder-decoder architecture, some having an additional belief tracker~\cite{wenN2N17} or latent intent variable~\cite{pmlr-v70-wen17a}. Outside of dialog management, RL-based sequence prediction algorithms have been proposed.  Specifically, \citet{ranzato2015sequence} and \citet{Bahdanau2017an} use policy
gradient and actor critic to learn to generate sequences for machine translation and document summarization. However, when directly applied to dialog generation, these methods would only optimize the reward of next utterance instead of
the overarching goal of the dialog. \citet{li2016deep} proposed a RL-based dialog generation algorithm which optimizes a set of rewards such as information flow and semantic coherence, focusing on open-domain dialogs. \citet{kandasamy2016batch}
proposed a batch policy gradient method that can train dialog policies offline from
existing dataset. Our proposed algorithm is a sequence prediction method, but differs from their work in that we do not assume the rewards are available in the dataset. Instead, we make the more realistic assumption that the trained agent utterances in the dataset are of high quality and can be treated as targets to learn, which is the case in TACT datasets.

\section{Method}
\subsection{Problem Formulation}
\label{sec:problem_formulation}
We represent each dialog in the training dataset as a sequence of pairs $D =
\{(x^k, y^k)\}_{k=1}^K$, where $K$ is the number of turns, $x^k$
is the user utterance at turn $k$, and $y^k$ is the agent utterance at turn
$k$. We define the context of the dialog at turn $k$ as $c^k = (x^1, y^1, ...,
y^{k-1}, x^k)$, that is, the concatenation of all utterances before
$y^k$. Usually, in a goal-oriented dialog dataset, there are API calls which are
issued by agents to external systems. We treat an API call as an agent utterance
starts with a special {\tt api\_call} token followed by the parameters of that API
call.

Given the context $c^k$ of the dialog, the algorithm generates an agent
utterance, token by token. We formulate this process as a Markov decision
process (MDP). The action space $\mathcal{A}$ of this MDP is the agent's
vocabulary. At each time $t$, an action is performed, that is, a token from the
vocabulary is generated. Let $z_{t-1}^k = (a^k_1, a^k_2, ..., a^k_{t-1})$ be the
sequence of tokens generated until time $t-1$. Then, the state of the MDP is
defined as $s^k_t = (c^k, z^k_{t-1})$, that is, the concatenation of the dialog
context and the tokens that have already generated. The stochastic policy
$\pi_{\theta}(\cdot|s^k_t)$ produces a distribution over the tokens in the vocabulary
based on the current state, where $\theta$ is the parameters of the policy we
try to learn. One token $a^k_t \sim \pi(\cdot|s^k_t)$ is sampled
from the distribution, and the state is deterministically transitioning to
$s^k_{t+1} = (c^k, z^k_{t})$. The terminal states are the states in which the
last generated token is a special EOS (end-of-sentence) token. We use $z^k$ to
denote the generated agent utterance, which has a EOS token at the end. The
length of $z^k$ is denoted by $T$. We will shortly
describe how we define the reward function of this MDP. The goal of the learning
algorithm is to maximize the cumulative rewards of the generated sequences,
which is defined in section \ref{sec:onpolicy}.

\subsection{Algorithm}
We parameterize our stochastic policy $\pi_{\theta}(\cdot|s_t^k)$ as an attention-based sequence-to-sequence (\textsc{Seq2Seq}) model~\cite{bahdanau2015neural}. Given a state, $s^k_t$, the encoder reads the dialog context, $c^k$, and the decoder outputs a probability distribution over the vocabulary conditioned on the last generated token, $a_{t-1}^k$, and the RNN hidden states of both $c^k$ and $z^{k}_{t-1}$. Note our algorithm is agnostic to the choice of the underlying model used to parameterize the policy; thus, state-of-the-art
encoder-decoder techniques, such as different attention mechanisms, can be easily
applied here to improve the model.

One natural choice to optimize our model parameters is to use \textit{on-policy}
policy gradient~\cite{williams1992simple}. In this setting, the algorithm
explores the action space $\mathcal{A}$ and generates an agent utterance $z^k$,
then $z^k$ is scored by a pre-defined reward function $r(z^k, D)$. One problem
with this is that the action space $\mathcal{A}$ is the agent token vocabulary, which 
is quite large and extremely difficult to effectively explore in order to
learn a good policy. To solve this problem, \citet{ranzato2015sequence} and
\citet{Bahdanau2017an} proposed to start from optimizing cross-entropy loss, and
then slowly deviate from it to let the model explore.

In this paper, we propose to combine \textit{on-policy} policy gradient with
\textit{off-policy} policy gradient~\cite{degris2012off,kandasamy2016batch}.
We show that if the actions in the dataset have high returns (future cumulative rewards),
such as the actions in TACT datasets, then off-policy policy
gradient can further speed up the convergence of the overall
algorithm. Moreover, off-policy learning enables us to utilize any existing rewards in
the dataset.

\subsubsection{Learning with On-policy Policy Gradient}
\label{sec:onpolicy}
Since the transition function of the defined MDP is known (i.e., an action that
adds a word to an utterance will always transition to the state where the word
is concatenated to the said utterance), once we define a
reward function $r(z^k, D)$ based on the generated utterance $z^k$ and the
dialog $D$ in the dataset, we can use on-policy policy gradient to learn the
optimal policy.

We define two types of rewards: \textit{utterance-level} rewards (i.e., rewards distributed within a single utterance) and
\textit{dialog-level} rewards (i.e., rewards that cross single utterance boundaries). Utterance-level rewards capture the quality of
the generated agent utterances $z^k$ compared with the existing agent utterances
$y^k$ in the training data. Example reward functions can include the semantic distance between the two utterances or simply the cosine similarity. In our experiments, we use BLEU scores~\cite{papineni2002bleu} to derive utterance-level rewards, which is one of the most popular metrics in machine translation and has been shown to correlate well with human judgement.

The dialog-level rewards, on the other hand, captures the contribution of the generated
utterances to achieving the dialog goals. We observe that goal-oriented dialogs
are usually driven by issuing API calls to a database or knowledge base. For
example, in a scenario in which a customer wants to return a product bought from
a shopping website, a customer service agent has to first issue an API call to
pull out the customer's profile. Then, the agent needs to check whether the
customer is eligible for a refund. Finally, the agent issues another API call to
start the refund process, and the goal of the dialog is achieved. API calls
usually have parameters. In the above example, the parameters for the first API
call could be the customer's email address, and the parameters for the second
one could be the order id and produce id. API calls are usually logged and
available for training.

It is critical for a policy to predict API calls and the parameters of the
API calls correctly. When a API call is issued later than the one in the
training data, it increases the number of turns of the dialog and may decrease
the user experience. A worse case is when a API call is issued earlier than the
one in the training data, since it has a risk to guess the parameters of
the API call. Worst of all is the case when the API call is never issued and the
goal of the dialog is missed. As a results, the dialog-level reward function we
define gives different negative rewards for API calls that are made too late or
too early, and a positive reward for each correct parameter of a API call
that is made on time.

We define the reward of a generated agent
utterance $z^k$ as the sum of utterance-level and dialog-level rewards:
\begin{align}
  \hspace{-0.2cm}
  \label{eq:reward_func}
  r(z^k, D) = \text{BLEU}(z^k, y^k) +
    \begin{cases}
     0 & \quad \text{if } z^k \text{ and } y^k \text{ are not API calls} \\
     -\lambda_a & \quad \text{if }  z^k \text{ is not an API call but } y^k \text{ is} \\
     -\lambda_b & \quad \text{if }  z^k \text{ is an API call and } k < k' \\
     -\lambda_c & \quad \text{if }  z^k \text{ is an API call and } k > k' \\
     \lambda_d * \text{\#correct\_parameters} & \quad \text{if }  z^k \text{ is
       an API call and } k = k' 
    \end{cases}
\end{align}
where $\lambda_a $, $\lambda_b$, $\lambda_c$, and $\lambda_d$ are
hyper-parameters of the algorithm 
and are between $0$ and $1$, and $k'$ is the actual turn in which the
corresponding API call was issued in the training data. BLEU scores are between 0 and 1.

The reward $r(z^k, D)$ is assigned to the last action in $z^k$ (an EOS token)
and all the intermediate actions get a reward of $0$. As a result, the reward
signals are sparse and the learning process can be slow. Similar to
~\citet{Bahdanau2017an}, we use reward shaping~\cite{daishi1999policy} to deal
with this problem. We define the potential function $\Phi(z^k_t) =
\text{BLEU}(z^k_t, y^k)$ for incomplete utterances and $\Phi(z^k) = 0$ for
complete ones. Then the reward for action $a^k_t$ is $r^k_t = \Phi(z^k_t) -
\Phi(z^k_{t-1})$ for all $t < T$, and the reward for the last action $a^k_T$ is
$r^k_T = r(z^k, D)$. The optimal policy doesn't change under reward shaping.

The objective function to maximize is
\begin{align*}
  J(\theta) = \mathbb{E}\left\{ 
  \sum_{t=1}^T \gamma^{t-1} r^k_t \ \vert \  \pi_{\theta} \right\}
\end{align*}
where $\gamma$ is the discount factor.
The policy gradient is estimated by~\cite{sutton2000policy}:
\begin{align}
  \label{eq:jonpolicy}
  \nabla J_{\text{on-policy}}(\theta) = 
  \sum_{t=1}^T
  \left[ \nabla_{\theta} \log \pi_{\theta}(a^k_t|s^k_t) \left(Q^{\pi} (s^k_t, a^k_t) - b(s^k_t)\right) \right]
\end{align}
where $b(s^k_t)$ is called the baseline~\cite{williams1992simple}, which is the
average return of the current policy at $s^k_t$.
We can use a function approximator such as another neural network to estimate
$Q^{\pi}(s^k_t, a^k_t)$. Then the algorithm falls into the actor-critic
framework~\cite{Bahdanau2017an}. However, for simplicity, in this paper we use
Monte Carlo estimate of
$Q^{\pi}(s^k_t, a^k_t)$, which is similar to the REINFORCE~\cite{williams1992simple}
algorithm:
\begin{align*}
  Q^{\pi}(s^k_t, a^k_t) = \sum_{t'=t}^T \gamma^{t'-t} r^k_{t'}
  \end{align*}

\subsubsection{Acceleration with Off-Policy Policy Gradient}
When the action and state space are large, on-policy RL algorithms can converge
slowly. One reason is that on-policy algorithms have to explore and experience actions with high returns before it can learn to pick these
good actions. However, in our case, the number of all possible sequences to
generate is $|\mathcal{A}|^{T}$, where $|\mathcal{A}|$ is the size of the
agent token vocabulary and on-policy exploration is difficult.
Meanwhile, in TACT datasets, actions are generated by trained
agents and are expected to have high returns. As the dialogs in these datasets are
demonstrations of good trajectories, it is sensible to reduce exploration efforts by using this information.

In this paper, we propose to use off-policy policy gradient~\cite{degris2012off}
to exploit such information. Off-policy policy gradient essentially maximize the
probability of actions in the dataset, weighted by importance sampling ratios
and the returns of these actions. Denoting the $i$-th token in $y^k$ as $o_i^k$ and $\tilde{T}$ as the length of $y^k$, the state of the MDP is represented by
$\tilde{s}^k_t = (c^k_t, o_1^k, ..., o_{t-1}^k)$, similar to the on-policy setting.
Off-policy policy gradient is estimated by~\cite{kandasamy2016batch}:
\begin{align}
  \label{eq:joffpolicy}
  \nabla J_{\text{off-policy}}(\theta) = 
  \sum_{t=1}^{\tilde{T}}
  \left[ \frac{\pi_{\theta}(o_t^k|\tilde{s}_t^k)}{q(o_t^k|\tilde{s}_t^k)} \nabla_{\theta} \log \pi_{\theta}(o^k_t|\tilde{s}^k_t) \left(Q^{\pi} (\tilde{s}^k_t, o^k_t) - b(\tilde{s}^k_t)\right) \right]
\end{align}

where
\begin{align}
  \label{eq:qso}
  Q^{\pi}(\tilde{s}^k_t, o^k_t) = \left[ \prod_{t'=t}^{\tilde{T}} \frac{\pi_{\theta}(o^k_t|\tilde{s}^k_t)}{q(o^k_t|\tilde{s}^k_t)} \right] \left[ \sum_{t'=t}^{\tilde{T}} \gamma^{t'-t} \tilde{r}^k_t\right]
\end{align}
and $q(\cdot|\tilde{s}_t^k)$ is the behavior policy, that is, the policy used
to generate the training dataset. We use the same reward function defined in
section \ref{sec:onpolicy} to calculate $\tilde{r}^k_t$. However, if there
already exists reward signals in the training data, that can also be used as
$\tilde{r}^k_t$.

Note that in equation (\ref{eq:joffpolicy}) and (\ref{eq:qso}),
$\frac{\pi_{\theta}(o^k_t|\tilde{s}^k_t)}{q(o^k_t|\tilde{s}^k_t)}$ is called the
importance sampling coefficient. We need the behavior policy
$q(\cdot|\tilde{s}^k_t)$ to calculate this coefficient, however, the behavior
policy is usually unknown. To solve this problem, ~\citet{kandasamy2016batch}
proposed to train another RNN model to estimate $q(\cdot|s^k_t)$. However, the
importance sampling coefficient can be very large if $q(\cdot|s^k_t)$ is very
small, and this can lead to high variance of the gradients. To deal with this
problem, one can clip this value, or simply set it to a constant value if
$\pi_{\theta}(\cdot|s^k_t)$ is close to
$q(\cdot|s^k_t)$~\cite{ionides2008truncated,munos2016safe}. Note that if we set the
coefficient to a constant 
value of $1.0$ and only use utterance-level rewards, then off-policy policy
gradient reduces to optimizing cross-entropy loss -- meaning that the proposed algorithm will outperform the underlying encoder-decoder if a reasonable reward function can be defined.

We update the policy parameters with a convex combination of the on-policy and
off-policy gradient:
\begin{align*}
  \nabla J(\theta) = (1-\lambda_e)\nabla J_{\text{off-policy}}(\theta) +
  \lambda_e \nabla J_{\text{on-policy}}(\theta)
\end{align*}
where $\lambda_e \in [0, 1]$. Our algorithm is outlined in Algorithm \ref{algo:algo}.

\begin{algorithm}
  \DontPrintSemicolon
  \LinesNumbered
\caption{End-to-End Offline Dialog Policy Learning}
\label{algo:algo}
\KwIn{$\lambda_a$, $\lambda_b$, $\lambda_c$, $\lambda_d \in [0, 1]$: reward function
  hyper-parameters \newline
  $\lambda_e \in [0, 1]$: policy gradient coefficient}
\While{Not Converged}{
Sample a random dialog $D$ \;
\For{$k=1, ..., K$}{
Use the current policy $\pi_{\theta}$ to predict the agent utterance $z^k$ given
the context $c^k=(x^1, y^1, ..., y^{k-1}, x^k)$\;
Calculate $r(z^k, D)$ based on equation (\ref{eq:reward_func}).
Get $r(y^k, D)$ from the data if available, otherwise calculate $r(y^k, D)$ based on equation (\ref{eq:reward_func}) \;
Calculate $J_{\text{on-policy}}(\theta)$ and $J_{\text{off-policy}}(\theta)$ based on equation (\ref{eq:jonpolicy}) and (\ref{eq:joffpolicy}). \;
Update $\theta$ with $\nabla J(\theta) = (1-\lambda_e)J_{\text{off-policy}}(\theta) + \lambda_e J_{\text{on-policy}}(\theta) $\;
}
}
\end{algorithm}

\section{Experiments}
\subsection{Dataset}
We use the bAbI dialog task $6$ dataset~\cite{bordes2016learning} for our
experiments. The bAbI dialog task $6$ was converted from the $2$nd Dialog State
Tracking Challenge, and is in the context of restaurant search. The goal of
the dialog is to recommend a restaurant based on a user's preferences. In each
dialog, the agent asked the users questions about their preferences on type of
cuisine, location and price range. The users can also initiate the dialog by
providing these preferences. The agent then issued an API call to a knowledge
base, which returned a list of candidate restaurants and their attributes such as
rating, phone number, address, etc.\ The restaurant with the highest rating was
recommended to the user. The user can then ask further information such as the
phone number or address of the restaurants. The user may update their
preferences during the dialog, so there can be multiple API calls in one
dialog.
Although the agent utterances in the dataset are not generated by trained
human agents, they are generated by a well-performed hand-crafted dialog policy. Therefore, the agent
utterances can still be treated as ground truth for learning purposes.

The data contains the raw transcripts of the dialogs, which includes the agent
utterances (including API calls and parameters of the API calls), user
utterances, and knowledge base responses of the API calls. The
train/validation/test set contains $1618$, $500$, and $1117$ dialogs
respectively. The size of the vocabulary is $1229$. The average number of
utterances per dialog is $14$, and the average number of records returned from
API calls is $40$. Each record is a tuple consists of a restaurant name, an
attribute name, and an attribute value.

\subsection{Training}
One benefit of our algorithm is end-to-end training. In our experiments, we
trained on the raw data without any data pre-processing such as normalizing
tokens, replacing entities with special tokens, and similar procedures. We also fed all the knowledge base responses to the encoder, so the algorithm has to learn to pick the restaurant with the highest rating. Note that the knowledge base responses
are long lists of restaurants and their attributes, so this dramatically
increases the length of the dialog contexts (i.e.\ the length of the input
sequences of the encoder). The longest dialog contexts in the training data
contains $1556$ tokens, the average length of the contexts is $152.94$. The
length of the longest agent utterance in the training data is $29$, and the
average length of the agent utterances is $10.07$. The purpose of using raw data
is to evaluate the performance of the algorithm without any domain knowledge.

We used the same model hyper-parameters as described in \citet{eric2017copy}, and encoder-decoder based model, such
that we can directly compare with their results. The encoder was a single layer
bi-directional LSTM, and the decoder was a single layer LSTM. The word embedding size was set to $300$, and the number of units in the LSTM was set to $353$. The input dropout keep rate was set to 0.8. We used the Adam
optimizer~\cite{kingma2014adam} and set the learning rate to $10^{-3}$. We set the
importance sampling coefficient to a constant value of $1.0$, and set the maximum length of
decoding sequence to $35$. For the reward function, since we are evaluating the
algorithm with
an existing dataset instead of real humans, issuing API call too early and
too late are equally bad in terms of our metrics, so we set
$\lambda_a=\lambda_b=\lambda_c=\lambda_d = 0.1$. For the convex combination of on-policy
and off-policy gradient, we set $\lambda_e = 0.3$.
We selected the model that performed the best on the validation dataset, and reported
the performance of that model on the test dataset.
We report the following metrics: 1) Per Response Accuracy: the accuracy of
predicting the next agent utterance. The prediction is correct only if all the
predicted tokens match the corresponding ones in the dataset. 2) BLEU score of
the generated utterances. 3) Precision, Recall and $F_1$ score (micro-averaged)
of issuing API 
calls. A prediction is considered as true positive if an API call is correctly
predicted as an API call, even if the parameters of the API call are 
wrong. 4) API call Exact Match: within all the true positives of issuing API
calls, the accuracy of predicting every parameter correctly.

The baseline algorithm used in our experiment is an attention-based \textsc{Seq2Seq}
model~\cite{bahdanau2015neural}, which is trained using the same
hyper-parameters as our algorithm. We also include the performance of Memory
Network reported by \citet{bordes2016learning}, and the performance of
copy-augmented \textsc{Seq2Seq} model reported by \citet{eric2017copy}.

\begin{table}[]
\centering
\begin{tabular}{|l|c|c|c|c|c|c|}
\hline
\multicolumn{1}{|c|}{\multirow{2}{*}{Model}} & Per Response   & \multirow{2}{*}{BLEU} & \multicolumn{4}{c|}{API Call}                                     \\ \cline{4-7} 
\multicolumn{1}{|c|}{}                       & Accuracy       &                             & Precision      & Recall         & $F_1$ score     & Exact Match    \\ \hline
Memory Network                               & 41.10          & -                           & -              & -              & -              & -              \\ \hline
Attention \textsc{Seq2Seq}                            & 47.29          & 57.36                       & 33.69          & \textbf{84.83} & 48.23          & 67.28          \\ \hline
\citet{eric2017copy}                                         & 48.00          & 56.00                       & -              & -              & -              & -              \\ \hline
Our algorithm                                & \textbf{48.69} & \textbf{58.25}              & \textbf{35.22} & 81.34          & \textbf{49.16} & \textbf{76.95} \\ \hline
\end{tabular}
\caption{Evaluation on bAbI task $6$. A dash indicates that the result is
  unavailable.}
\label{table:exp}
\end{table}

\subsection{Results}
The experiment results are shown in Table \ref{table:exp}, where we observe that our algorithm improved the BLEU score by 1.55\% compared with the attention-based
\textsc{Seq2Seq} model and by $4.02\%$ compared with \citet{eric2017copy}. This shows that
our learned policy achieved better utterance-level performance. Furthermore, we can easily update the reward function to optimize other utterance-level metrics or a
combination of different metrics.
Our algorithm also improved the $F_1$ score of issuing API calls by $1.93\%$,
and significantly improved the accuracy of API calls' parameters (exact match)
by $14.37\%$ compared with the attention-based \textsc{Seq2Seq} model. As we mentioned before, goal-oriented dialogs are usually driven by the API calls. Accordingly, better performance on API call metrics implies a better goal-oriented policy. Finally, our algorithm improved the per-response accuracy by $2.96\%$ compared with attention-based \textsc{Seq2Seq} and by $1.44\%$ compared with \citet{eric2017copy}.

\section{Conclusion}
In this paper, we propose an RL-based algorithm to learn goal-oriented dialog
policy. The algorithm enables the following useful contributions: 1) we define a reward function such that our algorithm can optimize both utterance-level and
dialog-level metrics. 2) we learn RL-based dialog policies by fully utilizing
TACT datasets without interacting with real users. 3) we improve the sample
efficiency of on-policy policy gradient by incorporating off-policy policy
gradient. 4) we parameterize our policy with an encoder-decoder architecture,
which enables end-to-end learning with no domain-specific knowledge. Compared
with recently proposed methods, our algorithm achieved better performance on both utterance-level and dialog-level metrics on bAbI dialog dataset. Our algorithm excels in the scenario when there exists a large amount of TACTs, which is the common case in industry setting. Therefore, we believe this work is an important step towards building industry-quality chatbots.

\small
\bibliography{bib}

\begin{thebibliography}{30}
\providecommand{\natexlab}[1]{#1}
\providecommand{\url}[1]{\texttt{#1}}
\expandafter\ifx\csname urlstyle\endcsname\relax
  \providecommand{\doi}[1]{doi: #1}\else
  \providecommand{\doi}{doi: \begingroup \urlstyle{rm}\Url}\fi

\bibitem[Bahdanau et~al.(2015)Bahdanau, Cho, and Bengio]{bahdanau2015neural}
Dzmitry Bahdanau, Kyunghyun Cho, and Yoshua Bengio.
\newblock Neural machine translation by jointly learning to align and
  translate.
\newblock In \emph{International Conference on Learning Representations}, 2015.

\bibitem[Bahdanau et~al.(2017)Bahdanau, Brakel, Xu, Goyal, Lowe, Pineau,
  Courville, and Bengio]{Bahdanau2017an}
Dzmitry Bahdanau, Philemon Brakel, Kelvin Xu, Anirudh Goyal, Ryan Lowe, Joelle
  Pineau, Aaron Courville, and Yoshua Bengio.
\newblock An actor-critic algorithm for sequence prediction.
\newblock In \emph{International Conference on Learning Representations}, 2017.

\bibitem[Bordes and Weston(2017)]{bordes2016learning}
Antoine Bordes and Jason Weston.
\newblock Learning end-to-end goal-oriented dialog.
\newblock In \emph{International Conference on Learning Representations}, 2017.

\bibitem[Degris et~al.(2012)Degris, White, and Sutton]{degris2012off}
Thomas Degris, Martha White, and Richard~S Sutton.
\newblock Off-policy actor-critic.
\newblock In \emph{Proceedings of the 29th International Coference on
  International Conference on Machine Learning}, pages 179--186, 2012.

\bibitem[Dhingra et~al.(2017)Dhingra, Li, Li, Gao, Chen, Ahmed, and
  Deng]{dhingra2017}
Bhuwan Dhingra, Lihong Li, Xiujun Li, Jianfeng Gao, Yun{-}Nung Chen, Faisal
  Ahmed, and Li~Deng.
\newblock Towards end-to-end reinforcement learning of dialogue agents for
  information access.
\newblock In \emph{Proceedings of the 55th Annual Meeting of the Association
  for Computational Linguistics}, pages 484--495, 2017.

\bibitem[Eric and Manning(2017)]{eric2017copy}
Mihail Eric and Christopher~D Manning.
\newblock A copy-augmented sequence-to-sequence architecture gives good
  performance on task-oriented dialogue.
\newblock In \emph{Proceedings of the 15th Conference of the European Chapter
  of the Association for Computational Linguistics}, page 468, 2017.

\bibitem[Ionides(2008)]{ionides2008truncated}
Edward~L Ionides.
\newblock Truncated importance sampling.
\newblock \emph{Journal of Computational and Graphical Statistics}, 17\penalty0
  (2):\penalty0 295--311, 2008.

\bibitem[Kandasamy et~al.(2017)Kandasamy, Bachrach, Tomioka, Tarlow, and
  Carter]{kandasamy2016batch}
Kirthevasan Kandasamy, Yoram Bachrach, Ryota Tomioka, Daniel Tarlow, and David
  Carter.
\newblock Batch policy gradient methods for improving neural conversation
  models.
\newblock In \emph{International Conference on Learning Representations}, 2017.

\bibitem[Kingma and Ba(2014)]{kingma2014adam}
Diederik Kingma and Jimmy Ba.
\newblock Adam: A method for stochastic optimization.
\newblock \emph{arXiv preprint arXiv:1412.6980}, 2014.

\bibitem[Li et~al.(2016)Li, Monroe, Ritter, Jurafsky, Galley, and
  Gao]{li2016deep}
Jiwei Li, Will Monroe, Alan Ritter, Dan Jurafsky, Michel Galley, and Jianfeng
  Gao.
\newblock Deep reinforcement learning for dialogue generation.
\newblock In \emph{Proceedings of the 2016 Conference on Empirical Methods in
  Natural Language Processing}, pages 1192--1202, 2016.

\bibitem[Li et~al.(2014)Li, He, and Williams]{li2014temporal}
Lihong Li, He~He, and Jason~D Williams.
\newblock Temporal supervised learning for inferring a dialog policy from
  example conversations.
\newblock In \emph{Spoken Language Technology Workshop (SLT), 2014 IEEE}, pages
  312--317. IEEE, 2014.

\bibitem[Munos et~al.(2016)Munos, Stepleton, Harutyunyan, and
  Bellemare]{munos2016safe}
R{\'e}mi Munos, Tom Stepleton, Anna Harutyunyan, and Marc Bellemare.
\newblock Safe and efficient off-policy reinforcement learning.
\newblock In \emph{Advances in Neural Information Processing Systems}, pages
  1054--1062, 2016.

\bibitem[Ng et~al.(1999)Ng, Harada, and Russell]{daishi1999policy}
Andrew~Y Ng, Daishi Harada, and Stuart Russell.
\newblock Policy invariance under reward transformations: Theory and
  application to reward shaping.
\newblock In \emph{Proceedings of the 16th International Conference on Machine
  Learning}, 1999.

\bibitem[Papineni et~al.(2002)Papineni, Roukos, Ward, and
  Zhu]{papineni2002bleu}
Kishore Papineni, Salim Roukos, Todd Ward, and Wei-Jing Zhu.
\newblock Bleu: a method for automatic evaluation of machine translation.
\newblock In \emph{Proceedings of the 40th annual meeting on association for
  computational linguistics}, pages 311--318. Association for Computational
  Linguistics, 2002.

\bibitem[Peng et~al.(2017)Peng, Li, Li, Gao, Celikyilmaz, Lee, and
  Wong]{peng2017composite}
Baolin Peng, Xiujun Li, Lihong Li, Jianfeng Gao, Asli Celikyilmaz, Sungjin Lee,
  and Kam-Fai Wong.
\newblock Composite task-completion dialogue policy learning via hierarchical
  deep reinforcement learning.
\newblock In \emph{Proceedings of the 2017 Conference on Empirical Methods in
  Natural Language Processing}, pages 2221--2230, 2017.

\bibitem[Pietquin et~al.(2011)Pietquin, Geist, Chandramohan, and
  Frezza-Buet]{pietquin2011sample}
Olivier Pietquin, Matthieu Geist, Senthilkumar Chandramohan, and Herv{\'e}
  Frezza-Buet.
\newblock Sample-efficient batch reinforcement learning for dialogue management
  optimization.
\newblock \emph{ACM Transactions on Speech and Language Processing (TSLP)},
  7\penalty0 (3):\penalty0 7, 2011.

\bibitem[Ranzato et~al.(2016)Ranzato, Chopra, Auli, and
  Zaremba]{ranzato2015sequence}
Marc'Aurelio Ranzato, Sumit Chopra, Michael Auli, and Wojciech Zaremba.
\newblock Sequence level training with recurrent neural networks.
\newblock In \emph{International Conference on Learning Representations}, 2016.

\bibitem[Ross et~al.(2011)Ross, Gordon, and Bagnell]{ross2011reduction}
St{\'e}phane Ross, Geoffrey~J Gordon, and Drew Bagnell.
\newblock A reduction of imitation learning and structured prediction to
  no-regret online learning.
\newblock In \emph{International Conference on Artificial Intelligence and
  Statistics}, pages 627--635, 2011.

\bibitem[Serban et~al.(2016)Serban, Sordoni, Bengio, Courville, and
  Pineau]{Serban2016}
Iulian~V. Serban, Alessandro Sordoni, Yoshua Bengio, Aaron Courville, and
  Joelle Pineau.
\newblock Building end-to-end dialogue systems using generative hierarchical
  neural network models.
\newblock In \emph{Proceedings of the Thirtieth AAAI Conference on Artificial
  Intelligence}, pages 3776--3783, 2016.

\bibitem[Sordoni et~al.(2015)Sordoni, Galley, Auli, Brockett, Ji, Mitchell,
  Nie, Gao, and Dolan]{sordoni2015a}
Alessandro Sordoni, Michel Galley, Michael Auli, Chris Brockett, Yangfeng Ji,
  Margaret Mitchell, Jian-Yun Nie, Jianfeng Gao, and Bill Dolan.
\newblock A neural network approach to context-sensitive generation of
  conversational responses.
\newblock In \emph{Proceedings of the 2015 Conference of the North American
  Chapter of the Association for Computational Linguistics}, pages 196--205,
  2015.

\bibitem[Su et~al.(2016{\natexlab{a}})Su, Gasic, Mrksic, Rojas-Barahona, Ultes,
  Vandyke, Wen, and Young]{su2016continuously}
Pei-Hao Su, Milica Gasic, Nikola Mrksic, Lina Rojas-Barahona, Stefan Ultes,
  David Vandyke, Tsung-Hsien Wen, and Steve Young.
\newblock Continuously learning neural dialogue management.
\newblock \emph{arXiv preprint arXiv:1606.02689}, 2016{\natexlab{a}}.

\bibitem[Su et~al.(2016{\natexlab{b}})Su, Gasic, Mrk\v{s}i\'{c},
  Rojas~Barahona, Ultes, Vandyke, Wen, and Young]{su2016online}
Pei-Hao Su, Milica Gasic, Nikola Mrk\v{s}i\'{c}, Lina~M. Rojas~Barahona, Stefan
  Ultes, David Vandyke, Tsung-Hsien Wen, and Steve Young.
\newblock On-line active reward learning for policy optimisation in spoken
  dialogue systems.
\newblock In \emph{Proceedings of the 54th Annual Meeting of the Association
  for Computational Linguistics}, pages 2431--2441, 2016{\natexlab{b}}.

\bibitem[Su et~al.(2017)Su, Budzianowski, Ultes, Gasic, and
  Young]{su2017sample}
Pei-Hao Su, Pawe{\l} Budzianowski, Stefan Ultes, Milica Gasic, and Steve Young.
\newblock Sample-efficient actor-critic reinforcement learning with supervised
  data for dialogue management.
\newblock In \emph{Proceedings of the 18th Annual SIGdial Meeting on Discourse
  and Dialogue}, pages 147--157, 2017.

\bibitem[Sutton et~al.(2000)Sutton, McAllester, Singh, and
  Mansour]{sutton2000policy}
Richard~S Sutton, David~A McAllester, Satinder~P Singh, and Yishay Mansour.
\newblock Policy gradient methods for reinforcement learning with function
  approximation.
\newblock In \emph{Advances in neural information processing systems}, pages
  1057--1063, 2000.

\bibitem[Wen et~al.(2017{\natexlab{a}})Wen, Miao, Blunsom, and
  Young]{pmlr-v70-wen17a}
Tsung-Hsien Wen, Yishu Miao, Phil Blunsom, and Steve Young.
\newblock Latent intention dialogue models.
\newblock In \emph{Proceedings of the 34th International Conference on Machine
  Learning}, pages 3732--3741, 2017{\natexlab{a}}.

\bibitem[Wen et~al.(2017{\natexlab{b}})Wen, Vandyke, Mrk\v{s}i\'{c}, Gasic,
  Rojas~Barahona, Su, Ultes, and Young]{wenN2N17}
Tsung-Hsien Wen, David Vandyke, Nikola Mrk\v{s}i\'{c}, Milica Gasic, Lina~M.
  Rojas~Barahona, Pei-Hao Su, Stefan Ultes, and Steve Young.
\newblock A network-based end-to-end trainable task-oriented dialogue system.
\newblock In \emph{EACL}, pages 438--449, 2017{\natexlab{b}}.

\bibitem[Williams and Zweig(2016)]{williams2016end}
Jason~D Williams and Geoffrey Zweig.
\newblock End-to-end lstm-based dialog control optimized with supervised and
  reinforcement learning.
\newblock \emph{arXiv preprint arXiv:1606.01269}, 2016.

\bibitem[Williams et~al.(2017)Williams, Asadi, and Zweig]{P17-1062}
Jason~D Williams, Kavosh Asadi, and Geoffrey Zweig.
\newblock Hybrid code networks: practical and efficient end-to-end dialog
  control with supervised and reinforcement learning.
\newblock In \emph{Proceedings of the 55th Annual Meeting of the Association
  for Computational Linguistics}, pages 665--677, 2017.

\bibitem[Williams(1992)]{williams1992simple}
Ronald~J Williams.
\newblock Simple statistical gradient-following algorithms for connectionist
  reinforcement learning.
\newblock \emph{Machine learning}, 8\penalty0 (3-4):\penalty0 229--256, 1992.

\bibitem[Young et~al.(2013)Young, Ga{\v{s}}i{\'c}, Thomson, and
  Williams]{young2013pomdp}
Steve Young, Milica Ga{\v{s}}i{\'c}, Blaise Thomson, and Jason~D Williams.
\newblock Pomdp-based statistical spoken dialog systems: A review.
\newblock \emph{Proceedings of the IEEE}, 101\penalty0 (5):\penalty0
  1160--1179, 2013.

\end{thebibliography}
\bibliographystyle{plainnat}
\end{document}